\documentclass[acmtog]{acmart}
\renewcommand\footnotetextcopyrightpermission[1]{}
\usepackage{xcolor}
\usepackage{colortbl}
\usepackage{algorithm}
\usepackage{multirow}
\usepackage{algpseudocode} 
\usepackage{amsmath}
\definecolor{best}{RGB}{255, 128, 128}
\definecolor{second}{RGB}{255, 153, 51}
\definecolor{third}{RGB}{255, 243, 79}
\AtBeginDocument{%
  }

\acmJournal{TOG}

\acmSubmissionID{1635}

\begin{document}

\title{Gaussian Sculpting: End-to-End Controllable Surface Reconstruction via Field Optimization}


\author{Ke Jiaxin}
\affiliation{%
  \institution{Dalian University of Technology}
  \country{China}
}
\email{kjx20000616@mail.dlut.edu.cn}

\author{Juncheng Liu}
\affiliation{%
  \institution{Massey University}
  \country{New Zealand}
}
\email{jliu9@massey.ac.nz}

\author{Yi Wang}
\affiliation{%
  \institution{Dalian University of Technology}
  \country{China}
}
\email{dlutwangyi@dlut.edu.cn}
\orcid{0000-0002-2724-7834}
\thanks{Corresponding author.}

\author{Zhouhui Lian}
\affiliation{%
  \institution{Peking University}
  \country{China}
}
\email{lianzhouhui@pku.edu.cn}

\author{Bin Liu}
\affiliation{%
  \institution{Dalian University of Technology}
  \country{China}
}
\email{liubin@dlut.edu.cn}

\author{Shengfa Wang}
\affiliation{%
  \institution{Dalian University of Technology}
  \country{China}
}
\email{sfwang@dlut.edu.cn}

\author{Xiangjia He}
\affiliation{%
  \institution{School of Computer Science, University of Nottingham Ningbo China}
  \city{Ningbo}
  \country{China}
}
\email{Sean.He@nottingham.edu.cn}


\begin{abstract}
  3D Gaussian Splatting (3DGS) has recently enabled real-time novel view synthesis with impressive quality. However, it struggles to recover accurate surfaces under limited viewpoints and due to the inherent irregularity of Gaussian primitives. The resulting geometric errors are notoriously difficult to correct manually. To address these issues, we propose Gaussian Sculpting, a fully differentiable end-to-end framework for high-quality surface reconstruction. Our key insight is to anchor Gaussians onto an evolving differentiable surface, allowing them to guide signed distance field (SDF) optimization instead of extracting the surface only during post-processing. To enable stable gradient isolation during joint optimization, we design a bi-level training strategy in which the outer loop optimizes the geometry represented by the SDF, while the inner loop updates the Gaussians with the geometry fixed. We further impose constraints on Gaussian parameters to ensure consistency with the underlying surface, thereby improving both geometric and appearance fidelity during optimization. In addition, we introduce a multi-resolution subdivision scheme based on octree-like partitioning to preserve fine details while reducing memory consumption. Experiments on object-level scenes demonstrate that our method effectively removes redundant surfaces, recovers missing structures caused by limited viewpoints, and achieves strong reconstruction quality even at relatively low resolutions.
\end{abstract}



\begin{teaserfigure}
  \centering
  \includegraphics[width=\textwidth]{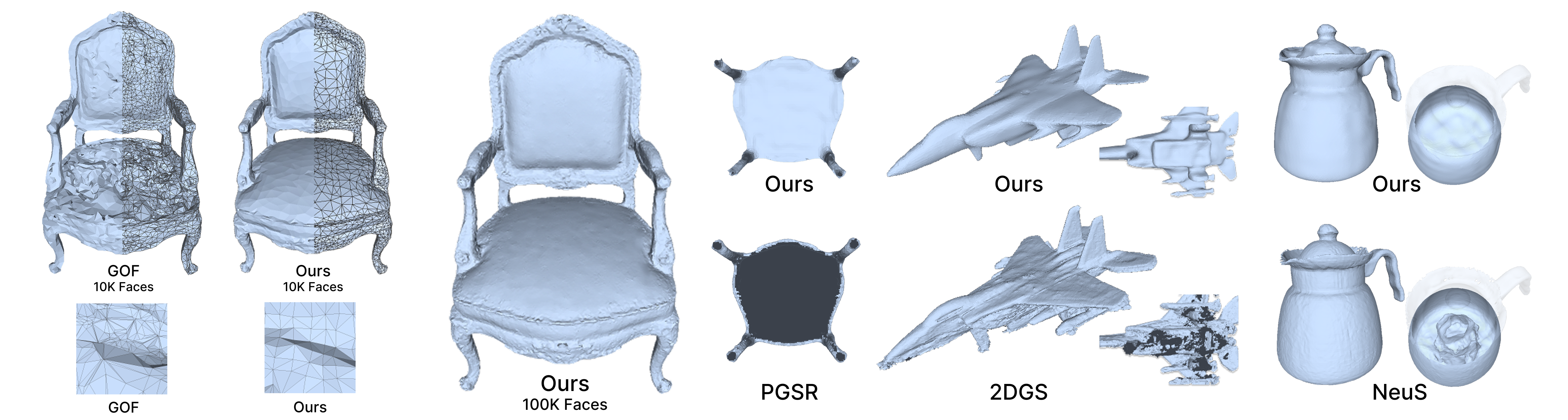}
  \caption{Gaussian Sculpting is an end-to-end surface reconstruction framework based on field optimization that progressively refines mesh geometry to recover fine details and eliminate erroneous structures. Compared with Gaussian opacity-field (GOF)~\cite{yu2024gaussian}, our method produces higher-quality meshes. We also alleviate the geometric incompleteness exhibited by PGSR~\cite{chen2024pgsr} and 2DGS~\cite{huang20242d}, and avoids the floating artifacts commonly produced by NeuS~\cite{wang2021neus}.}
  \label{fig:teaser}
\end{teaserfigure}
\maketitle

\section{Introduction}
Multi-view surface reconstruction remains a challenging problem in computer graphics and vision, holding significant value for applications such as digital products~\cite{chen2025meshanything,tang2023dreamgaussian,lu2025matrix3d,li2025caddreamer}, robotics~\cite{pan2025russo,fan2025neugrasp}, and augmented/virtual reality (AR/VR)~\cite{deng2022fov,ye2021superplane,zhang2021arcargo,pechko2025gs,kim2025play4d}. In recent years, 3D Gaussian Splatting (3DGS)~\cite{kerbl20233d} has become an effective method for novel view synthesis, due to its real-time rendering capability and efficient training. However, visually plausible renderings may still correspond to incomplete or unstable surfaces. Notably, downstream tasks such as game asset modeling and map construction demand more than just visual navigation. They require editable and structured geometric models. Thus, extracting meshes from Gaussian representations has emerged as a crucial research direction.

Prior methods for surface extraction from 3D Gaussians~\cite{guedon2024sugar,huang20242d,chen2024pgsr} treat reconstruction as a post-processing step, applying Marching Cubes (MC)~\cite{lorensen1998marching} or Poisson reconstruction~\cite{kazhdan2006poisson} to optimized Gaussians. Because this process is decoupled from end-to-end optimization, the resulting surface heavily depends on the Gaussian quality, yielding sensitivity to noise and poor geometric fidelity. Under limited or missing viewpoints, ambiguous Gaussian shapes and artifacts lead to incomplete geometry, which improvements in Gaussian geometry alone cannot resolve.
Recent field-optimized Gaussian methods~\cite{yu2024gsdf,yu2024gaussian} partially address these issues. However, due to the irregular and disordered nature of Gaussian distributions, neither the Gaussians nor their derived fields can faithfully align with a triangle mesh. This mismatch propagates discretization errors, including floating surfaces and sliver triangles~\cite{sliver2000}. Obtaining a watertight, clean, and continuously optimizable mesh from limited-view RGB images therefore remains a challenging open problem.

To address these limitations, we reformulate surface reconstruction by optimizing a signed distance field (SDF) instead of Gaussian primitives, leveraging differentiable Gaussian splatting for rendering supervision and integrating differentiable surface extraction, thereby establishing a fully end-to-end optimized pipeline.
Unlike prior differentiable surface extraction methods that initialize optimization with discrete, noisy explicit SDFs, we represent the field implicitly using a multi-layer perceptron (MLP), which provides continuous geometry, improved smoothness, and better control over resolution. During optimization, surfaces are extracted from the evolving field, and Gaussians are dynamically instantiated on the resulting triangle mesh for the splatting process.

Crucially, Gaussians are not treated as independent optimization variables for correcting geometry. Instead, they serve as geometry-aware rendering proxies that faithfully reflect the current surface quality and guide field optimization. To enforce this role, each Gaussian is tightly anchored to the mesh surface, with its position, orientation, and scale explicitly determined by the corresponding triangle vertices and normal vectors. Furthermore, we introduce a set of Gaussian constraints that regulate opacity, scale range, and spatial distribution, ensuring consistent alignment between Gaussian primitives and the underlying surface geometry while preventing rendering-induced artifacts.

To further stabilize optimization, we design a bi-level optimization framework in which the outer loop refines the SDF and surface geometry, while the inner loop updates Gaussian parameters with the geometry fixed. This strategy prevents rendering supervision from interfering with geometry optimization. Moreover, we adopt a progressive octree-like subdivision scheme that confines high-resolution computation to reliably reconstructed surface regions. As the geometry is progressively refined through field optimization under structured Gaussian supervision, we refer to our method as \textit{Gaussian Sculpting}.

Our main contributions are summarized as follows:

\begin{itemize}
    \item We propose \textbf{Gaussian Sculpting}, an end-to-end framework for surface reconstruction that uses constrained Gaussians to guide SDF optimization.

    \item We identify the inconsistency between Gaussian rendering and surface geometry, and introduce geometry and opacity constraints to better align Gaussians with the underlying surface.
    
    \item We develop a \textbf{bi-level optimization framework} with gradient isolation and progressive subdivision for stable geometry refinement.

    \item Our method reconstructs cleaner and more complete meshes than prior NeRF-based and Gaussian-based methods, reducing floating artifacts and recovering missing regions.
\end{itemize}

\begin{figure*}[t!] 
  \centering
  \includegraphics[width=1.0\linewidth]{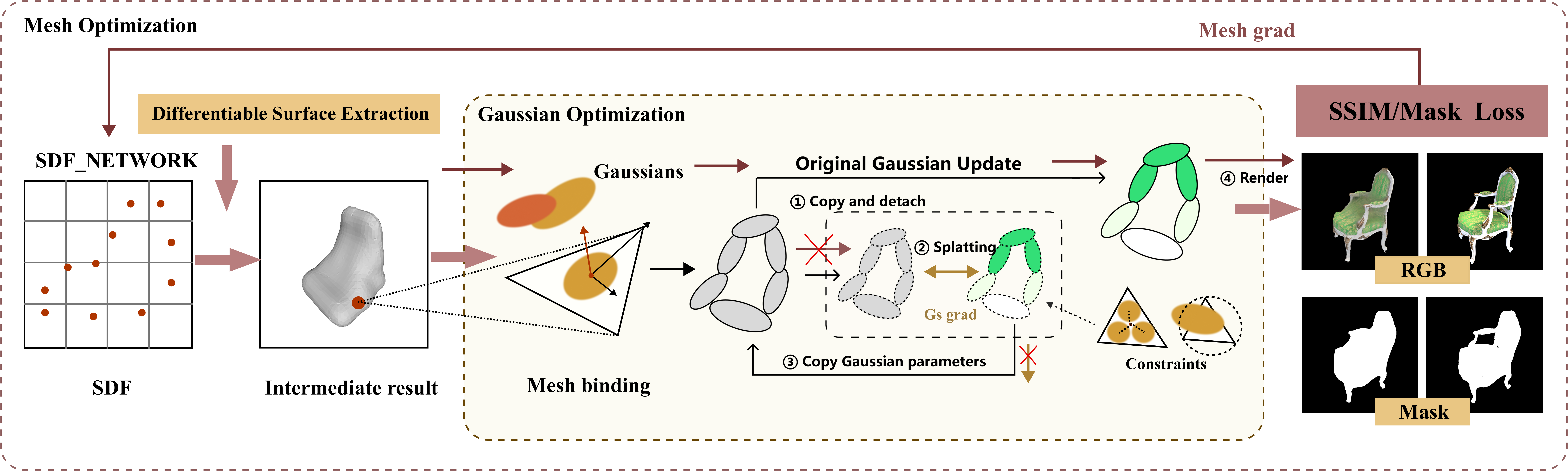} 
  \caption{Pipeline of Gaussian Sculpting. In each iteration of mesh optimization, the SDF network predicts SDF values at voxel vertices, from which an intermediate mesh is extracted through differentiable surface extraction. An inner loop independently optimizes Gaussians anchored to the mesh under Gaussian constraints. We use a detached representation to prevent rendering gradients from interfering with geometry optimization. Red-cross arrows mean those original Gaussians with SDF gradients are excluded from internal rendering. The optimized Gaussians are then used for rendering supervision with RGB and mask losses. Red arrows denote mesh-related gradients, while yellow arrows denote Gaussian-related gradients.} 
  \label{pipeline} 
\end{figure*}

\section{Related Work}

\subsection{Traditional Surface Reconstruction}
Traditional surface reconstruction aims to recover accurate geometric surfaces from sparse or noisy observations. Most methods follow a multi-view stereo (MVS) pipeline, establishing correspondences across multi-view images via patch matching~\cite{Bleyer2011PatchMatchS} and reconstructing surfaces through triangulation or implicit surface reconstruction~\cite{Implicit2025}.
The intermediate representations used in these methods include meshes, point clouds~\cite{furukawa2009accurate}, depth maps~\cite{campbell2008using}, and voxels~\cite{kutulakos2000theory}. These representations exhibit inherent limitations. Point-based and surface-based methods, such as Poisson Surface Reconstruction~\cite{kazhdan2006poisson}, rely heavily on accurate correspondence estimation. When texture is insufficient, reliable correspondence becomes difficult to establish, often leading to artifacts and missing regions.
Voxel-based approaches, such as those using MC~\cite{lorensen1998marching}, provide a structured representation but suffer from high memory consumption when applied to high-resolution reconstruction.

\subsection{Neural Surface Reconstruction}
With the advancement of neural networks, recent methods have explored learning-based surface reconstruction from single or multiple images in an end-to-end manner, representing shapes as point clouds~\cite{fan2017point,lin2018learning}, voxels~\cite{choy20163d,xie2019pix2vox}, or implicit fields~\cite{mescheder2019occupancy}.
Neural Radiance Fields (NeRF)~\cite{mildenhall2021nerf} synthesize photorealistic novel views via alpha compositing along camera rays but struggle to recover high-fidelity surface geometry. NeuS~\cite{wang2021neus} integrates signed distance functions into volume rendering, enabling more accurate surface reconstruction and improved robustness in scenes with sharp geometric transitions.
Despite their success, NeRF-based methods~\cite{mildenhall2021nerf,yariv2021volume,wang2021neus,wang2023neus2,yariv2023bakedsdf,li2023neuralangelo} are often limited by high inference cost and restricted representational capacity due to reliance on multi-layer perceptrons (MLPs). To address this, recent approaches decompose scene representation into points~\cite{xu2022point}, voxels~\cite{li2022vox}, and other explicit structures, reducing dependence on MLP-based implicit modeling.

\subsection{Gaussian Splatting based Surface Reconstruction} 
We categorize existing work into four directions:
(I) enhancing geometric adaptability, (II) extracting surfaces via post-processing, (III) improving Gaussian rendering, and (IV) leveraging Gaussians to guide geometry or field optimization.

2D Gaussian Splatting (2DGS)~\cite{huang20242d} collapses 3D Gaussian ellipsoids into oriented planar disks, yielding viewpoint-consistent geometry and a novel paradigm for fitting irregular Gaussians to smooth surfaces.  
Building on this, Gaussian Mesh Splatting (GaMeS) and Mani-gs~\cite{gao2025mani,waczynska2024games} redefine the parameters of Gaussians through geometric methods, introducing an adaptive triangle-aware Gaussian binding strategy. MeshGS~\cite{choi2024meshgs} binds Gaussians to surfaces via distance constraints, balancing alignment and detail for large-scale rendering.

Regarding surface extraction post-processing, related studies provide key information, including sample points, field distributions, and depth maps from Gaussian scenes, which serve as the basis for surface extraction. SuGar~\cite{guedon2024sugar} uses regularization to align Gaussians with surfaces and supports Poisson reconstruction via level-set sampling, but its externally defined SDF misrepresents surfaces and causes bubble artifacts. Methods exemplified by 2DGS~\cite{huang20242d} extract surfaces using MC~\cite{lorensen1998marching} or Marching Tetrahedra (MT)~\cite{lorensen1998marching} after TSDF fusion~\cite{izadi2011kinectfusion}. GOF~\cite{yu2024gaussian} replaces TSDF with opacity fields and MT, yet suffers high cost and poor mesh quality, failing to overcome Gaussian limitations.

For rendering quality optimization, PGSR~\cite{chen2024pgsr} introduces single‑view geometric and multi‑view photometric regularization terms within the Gaussian splatting framework to enhance reconstruction quality. GausSurf~\cite{wang2024gaussurf} guides 3D Gaussian optimization in texture‑rich regions through patch‑based multi‑view stereo matching, while in texture‑sparse regions it employs normal priors from pre‑trained models to constrain Gaussians. Although these methods improve the efficiency of Gaussian rendering and yield better results, they still fail to resolve surface quality issues.

Now several methods combine Gaussians with SDF optimization. GSDF~\cite{yu2024gsdf} jointly supervises rendering and reconstruction via mutual guidance; 3DGSR~\cite{lyu20243dgsr} loosely couples a neural SDF with 3DGS through geometric regularization; and GS-ROR2~\cite{zhu2025gs} enables bidirectional Gaussian-SDF refinement, pruning outliers and refining normals for reflective surfaces. Despite these advances, they remain constrained by noisy Gaussian estimates and inconsistent depth cues, which introduce floating artifacts and compromise geometric fidelity. Moreover, their typically decoupled training strategies weaken geometry-appearance interaction, limiting the effectiveness of joint optimization~\cite{yu2024gsdf}.

Motivated by these limitations, we propose a fully differentiable surface reconstruction framework that optimizes an SDF and enforces strict surface binding of Gaussian primitives.

\section{Method}
We propose an end-to-end SDF optimization framework for surface reconstruction from multi-view RGB images (Fig.~\ref{pipeline}). Our approach jointly optimizes an SDF network and Gaussian representations. Unlike methods that extract surfaces as a post-processing step, we employ a differentiable iso-surface extraction module to derive the surface from the current SDF. Gaussians are then instantiated on the extracted mesh and used for differentiable rendering. We further introduce Gaussian constraints to maintain consistency between Gaussian rendering and the underlying surface.
\subsection{Field Optimization}
This section introduces our implicit SDF formulation, the surface extraction process, and our surface-constrained Gaussian representation. 

\textbf{Implicit Field Representation:}
We represent the implicit SDF using an MLP that maps 3D coordinates to signed distance values and high-dimensional features.

\textbf{Differentiable Surface Extraction:} We extend Flexicubes~\cite{shen2023flexible} to support multi-resolution grids and SDF inputs, enabling differentiable iso-surface extraction with improved topological consistency and hierarchical adaptivity.
The extraction method introduces three groups of learnable parameters: interpolation weights $\alpha \in \mathbb{R}_{>0}^{8}$ and $\beta \in \mathbb{R}_{>0}^{12}$ per grid cell for dual vertex positioning, a split weight $\gamma \in \mathbb{R}_{>0}$ per grid cell for controlling adaptive triangulation, and a deformation vector $\delta \in \mathbb{R}^{3}$ per grid vertex for spatial alignment. These parameters enable gradient-based optimization while improving surface fitting quality.

\textbf{Surface Gaussian Model:}
Upon obtaining the reconstructed surface, we utilize Gaussian rendering to reflect its current quality. The original 3DGS relies on RGB supervision and often fails to establish a precise correspondence to the true object geometry. Therefore, we redesign the Gaussian parameterization. Our method initializes Gaussian attributes from mesh vertices and normals, and constrains them through color (opacity) contributions and geometric fitting (scale and distribution).

For a mesh with $F$ triangular faces, each face $\mathcal{T}_f$ is defined by three vertices $\{\mathbf{v}_{f,1}, \mathbf{v}_{f,2}, \mathbf{v}_{f,3}\} \subset \mathbb{R}^3$. We place $K$ Gaussians on each face. The center $\mathbf{m}_{f,k} \in \mathbb{R}^3$ of the $k$-th Gaussian on face $f$ is expressed as a convex combination of the face vertices:
\begin{equation}
\mathbf{m}_{f,k} = \sum_{i=1}^{3} \alpha_{f,k,i} \mathbf{v}_{f,i},
\label{eq:barycentric_center}
\end{equation}
where the barycentric weights $\alpha_{f,k,i}$ are learnable parameters satisfying $\alpha_{f,k,i} \ge 0$ and $\sum_{i=1}^3 \alpha_{f,k,i} = 1$.

The covariance matrix of each Gaussian on face $\mathcal{T}_f$ is parameterized as
$\Sigma_{f,k} = \mathbf{R}_{f} \mathbf{S}_{f,k} \mathbf{S}_{f,k}^\top \mathbf{R}_{f}^\top$.
The rotation matrix $\mathbf{R}_{f} = [\mathbf{r}_0, \mathbf{r}_1, \mathbf{r}_2]$ defines an orthonormal frame, where $\mathbf{r}_0$ is the unit normal of $\mathcal{T}_f$, $\mathbf{r}_1$ is a normalized edge direction of the face, and $\mathbf{r}_2$ is obtained by orthogonalizing the remaining edge direction with respect to $\mathbf{r}_0$ and $\mathbf{r}_1$.
The scaling matrix $\mathbf{S}_{f,k} = \mathrm{diag}(s_0, s_1, s_2)$ uses a small normal scale $s_0 = \varepsilon$ to flatten the Gaussian, and two tangential scales defined by:
\begin{equation}
s_1 = \tfrac{1}{2}\|\mathbf{v}_{f,2} - \mathbf{v}_{f,1}\|,\qquad 
s_2 = \tfrac{1}{2}\|\mathbf{v}_{f,3} - \mathbf{v}_{f,1}\|.
\label{eq:scaling}
\end{equation}
The factor $1/2$ provides a conservative initialization, preventing overly large Gaussians and excessive overlap during early optimization.

\textbf{Scale Constraint:} To ensure that the typical spatial extent of each Gaussian remains within the face-aligned minimum enclosing ellipse (MEE), we formulate the constraint as a spectral condition in the normalized space.
For face $\mathcal{T}_f$, the MEE is defined by semi-axes $s_1, s_2$ in the tangent frame. Let $D=\mathrm{diag}(s_1,s_2)$. The Gaussian covariance projected to the tangent plane is denoted as $\Sigma_k$.
We normalize the covariance as:
\begin{equation}
\tilde{\Sigma}_k = D^{-1} \Sigma_k D^{-1}.
\end{equation}
The Gaussian is fully contained~\cite{CALBERT20231958} in the MEE only if:
\begin{equation}
\lambda_{\max}(\tilde{\Sigma}_k) \le 1,
\end{equation}
where $\lambda_{\max}()$ denotes the largest eigenvalue of a matrix, corresponding to the maximum directional variance of the Gaussian.
We define the MEE constraint loss as:
\begin{equation}
\mathcal{L}_{\mathrm{MEE}} =
\max\left(0, \lambda_{\max}(\tilde{\Sigma}_k) - 1 \right).
\end{equation}

\textbf{Distribution Constraint:} To avoid degenerate Gaussian placements and encourage stable coverage over each triangular face, we introduce a distribution regularization on Gaussian centers in barycentric space.

I) \textit{Boundary avoidance loss}  
$\mathcal{L}_b$ discourages Gaussians from collapsing toward triangle edges:
\begin{equation}
\mathcal{L}_b = \frac{1}{F \cdot K \cdot 3} \sum_{f=1}^{F} \sum_{k=1}^{K} \sum_{i=1}^{3} \exp\bigl(-10\,\alpha_{f,k,i}\bigr).
\label{eq:Lb}
\end{equation}
The exponential form rapidly increases the penalty near triangle boundaries.

II) \textit{Distance constraint loss} $\mathcal{L}_d$ prevents excessive clustering between neighboring Gaussians by encouraging a target spacing $\delta_{\text{target}} = 1/\sqrt{K}$, which is empirically chosen to be proportional to the Gaussian density:
\begin{equation}
\mathcal{L}_d = \frac{1}{F \cdot K} \sum_{f=1}^{F} \sum_{k=1}^{K} \Bigl| \delta_{f,k} - \delta_{\text{target}} \Bigr|,
\label{eq:Ld}
\end{equation}
where the minimum distance for the Gaussian $k$ is
\begin{equation}
\delta_{f,k} = \min_{j \neq k} \sqrt{ \sum_{i=1}^{3} \bigl( \alpha_{f,k,i} - \alpha_{f,j,i} \bigr)^2 }.
\label{eq:min_distance}
\end{equation}

III) \textit{Coverage loss} $\mathcal{L}_c$ promotes that the Gaussians on a face span the full barycentric domain, the per‑coordinate range covers most of $[0,1]$. 
Let $r_{f,i} = \max_k \alpha_{f,k,i} - \min_k \alpha_{f,k,i}$ be the range of the barycentric coordinate for vertex $i$ on face $f$. 
We penalize cases where this range falls below a threshold $\tau = 0.8$:
\begin{equation}
\mathcal{L}_c = \frac{1}{F \cdot 3} \sum_{f=1}^{F} \sum_{i=1}^{3} \max\!\bigl(0,\; \tau - r_{f,i}\bigr).
\label{eq:Lc}
\end{equation}
The final loss $\mathcal{L}_{\text{Dist}}$ is defined as
\begin{equation}
\mathcal{L}_{\text{Dist}} = \lambda_b\,\mathcal{L}_b + \lambda_d\,\mathcal{L}_d + \lambda_c\,\mathcal{L}_c,
\label{eq:L_uniform}
\end{equation}
where $\lambda_b$, $\lambda_d$, and $\lambda_c$ are balancing weights that control the contribution of each loss term.

\textbf{Opacity Constraint:}
To maintain stable rendering supervision and prevent erroneous geometry from being hidden by low-opacity Gaussians, we remove the gradient of opacity and fix it close to 1:
\begin{equation}
O_i = \sigma^{-1}(\theta), \quad \theta \to 1,
\label{eq:opacity_constraint}
\end{equation}
where $\sigma^{-1}(\cdot)$ is the inverse sigmoid function, $O_i$ is the opacity parameter of the $i$-th Gaussian, and $\theta$ is a constant close to 1.

\subsection{Resolution Control}
\begin{figure}[tb]
\centering
\includegraphics[width=1.0\columnwidth]{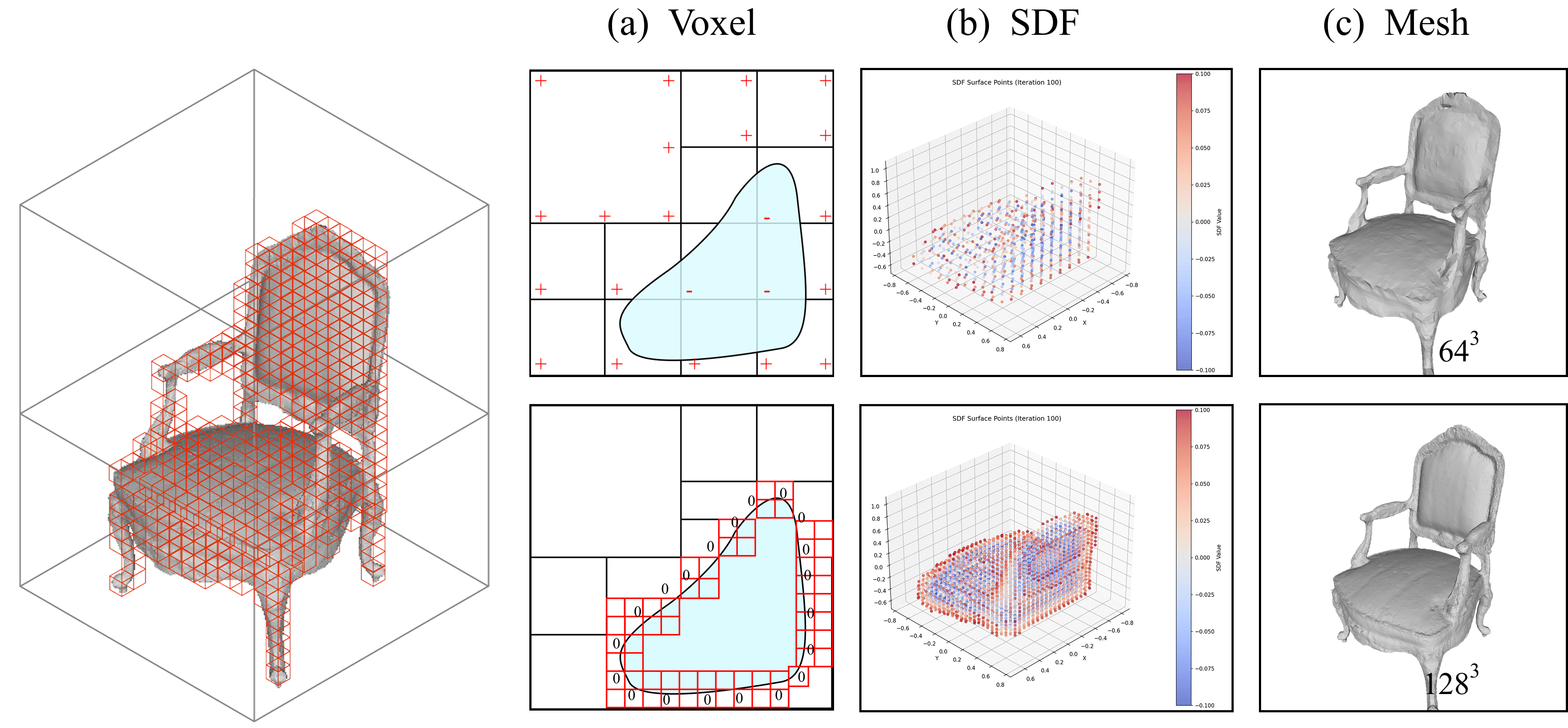}
\caption{Use an adaptive grid refinement method to control resolution. Left: 3D view of surface-only refinement. (a) Voxel subdivision: Targeted refinement of surface-intersecting voxels via SDF checks. (b) Refined SDF: Low-res (top) vs. high-res (bottom) fields. (c) Meshes: 64³ (top) vs. 128³ (bottom) reconstructions.}
\label{fig:resolution}
\end{figure}
During initialization, we voxelize the scene at the target resolution and evaluate the SDF field. Since object surfaces occupy only a small portion of the volume, uniformly high-resolution optimization leads to substantial computational overhead. We therefore adopt an octree-like adaptive subdivision strategy~\cite{frisken2000adaptively}, using coarse voxels in empty regions and subdividing surface-related voxels into finer sub-voxels for detail reconstruction (Fig.~\ref{fig:resolution}). To preserve correct dual-vertex computation, adjacent voxels are subdivided consistently. We further support local refinement within user-specified bounding regions to reduce early-stage subdivision overhead. Additional details on refinement region selection and vertex remapping are provided in the supplementary materials.
\subsection{Training}
To prevent unreliable gradients from contaminating geometry optimization, we formulate training as a bi-level optimization problem. The outer loop optimizes the surface geometry, while the inner loop only optimizes a Gaussian scene conditioned on the current surface. At each outer iteration, we initialize a Gaussian scene $G$ from the surface as described in Section~3.1, which remains gradient-connected to the surface parameters.
We then construct a detached copy $\tilde{G}$ for inner-loop optimization. The inner loop trains $\tilde{G}$ using several steps of gradient descent under the same photometric reconstruction objective, allowing the Gaussian representation to adapt to the current surface geometry.
The photometric loss~\cite{SSIM2004} is defined as:
\begin{equation}
\mathcal{L}_{\text{RGB}} = (1-\lambda)\mathcal{L}_1 + \lambda \mathcal{L}_{\text{D-SSIM}},
\label{eq:photo_loss}
\end{equation}
computed between rendered and ground-truth images.

After inner optimization, we obtain $\tilde{G}^*$, whose parameters are copied back to the gradient-connected Gaussian scene $G$. The outer loop then evaluates the same photometric loss on $G$, enabling gradients to update the surface geometry. To stabilize early optimization, we additionally apply a silhouette mask loss before switching to photometric supervision.
\begin{align}
\tilde{G}^* &= \arg\min_{\tilde{G}} \mathcal{L}_{\text{RGB}}(S, \tilde{G}), \\
S^* &= \arg\min_{S} \mathcal{L}_{\text{RGB}}(S, G),
\end{align}
where the inner loop optimizes $\tilde{G}$ under fixed surface geometry $S$, and the outer loop updates $S$ using the synchronized Gaussian scene $G$. For implementation details of the training framework and gradient isolation, please refer to the supplementary materials.

\section{Experiment}
\begin{table*}[htbp]
  \centering
  \caption{The 12 objects are shown from left to right in groups of four, following the dataset's native difficulty grouping with increasing difficulty. Our method accurately and robustly reconstructs all objects, while others fail or have large errors on a few objects. ``—'' denotes reconstruction failure.}
  \resizebox{\textwidth}{!}{%
  \begin{tabular}{l|cccc|cccc|cccc|c}
  \toprule
     CD ($10^{-3}$) $\downarrow$ & Banana & Dumpling & Walnut & Peanut & Teapot & Toy\_plane & Handbag & Pumpkin & Toy\_train & Durian & Kennel & Pan & Mean \\
  \midrule
    NeRF & 38.22&68.54 &252.81 &69.43 &65.96 & 13.76&70.24 & 239.94& 46.75&25.51 &\cellcolor{third}13.36 &13.20 &76.48 \\
    NeuS & 5.95& 27.70&51.02 &12.68 &29.13 &\cellcolor{best}5.51 &— &44.05 &\cellcolor{second}8.87 & 19.24& \cellcolor{second}7.12&11.07 &20.21 \\
  \midrule
    SuGar & 74.18 & 67.16 & 17.99 & 21.33 & 75.95 & 95.78 & 19.33 & 58.50 & 80.26 & 111.20 & 36.90 & 31.28 & 57.49 \\
    2DGS & 11.70 & 6.88 & 52.37 & 29.26 & 36.75 & 15.63 & 36.63 & 71.45 & 29.96 & 25.05 & 27.55 & 9.18 & 29.37 \\
    PGSR & \cellcolor{second}7.24 & 15.04 & \cellcolor{third}8.62 & 10.22 & 17.02 & 15.67 & \cellcolor{second}11.67& \cellcolor{third}30.87 & 11.77 & 65.99 & 13.49 & \cellcolor{third}9.15 & 18.06 \\
    GOF & \cellcolor{third}9.10 & 4.95 & \cellcolor{second}8.31 & \cellcolor{best}6.24 & \cellcolor{second}6.39 & \cellcolor{second}7.48 & \cellcolor{best}7.41 & 72.17 & 12.35 & \cellcolor{second}22.37 & 15.52 & \cellcolor{second}6.60 & \cellcolor{third}14.91 \\
  \midrule
    GSDF &1.60 & 6.64 & 4.72 &\cellcolor{second}6.76 & \cellcolor{third}8.14 &— &— & \cellcolor{second}9.71 & \cellcolor{third}10.56 & \cellcolor{third}24.77 & 59.18&—& \cellcolor{second}14.68 \\
    \textbf{Ours} & \cellcolor{best}4.78 & 11.66 & \cellcolor{best}6.17 & \cellcolor{third}7.35 & \cellcolor{best}6.10 & \cellcolor{third}7.83 & \cellcolor{third}14.38 & \cellcolor{best}8.73 & \cellcolor{best}8.54 & \cellcolor{best}21.71 & \cellcolor{best}5.90 & \cellcolor{best}5.95 & \cellcolor{best}9.09 \\
  \bottomrule
  \end{tabular}%
  }
  \label{tab:Omini_results}
\end{table*}
\subsection{Experiment Settings}
\textbf{Details:}
We employ an MLP network with 256 hidden units and 8 fully connected layers as the implicit representation for the SDF, incorporating skip connections, positional encoding, geometric initialization, and weight normalization. The learning rate for the SDF network is set to 0.0001. The inner Gaussian training follows the default settings of 3DGS~\cite{kerbl20233d}, with 1000 iterations per internal step. All experiments are conducted on an RTX 3090 GPU with 24 GB of VRAM. 

 \textit{Training time.} The initial surface optimization converges within a few minutes, while most computational cost comes from inner-loop Gaussian refinement. The training time for a single scene typically ranges from 6 to 18 hours, depending on scene complexity and number of optimization iterations.

\textbf{Baselines:}
We select three categories of baselines for comparison: (I) NeRF and its SDF-based extensions~\cite{mildenhall2021nerf,wang2021neus,yariv2021volume}, (II) 3D Gaussian Splatting (3DGS)–based reconstruction methods~\cite{gao2024relightable,guedon2024sugar,huang20242d,chen2024pgsr,yu2024gaussian}, and (III) joint approaches where 3DGS is used to guide SDF optimization~\cite{lyu20243dgsr,yu2024gsdf}.
This selection enables a systematic evaluation of neural surface and Gaussian-based methods in isolation, and existing strategies for combining the two representations.

\textbf{Datasets and Evaluation Metrics:}
We adopt the NeRF Synthetic~\cite{mildenhall2021nerf} and OmniObject3D~\cite{wu2023omniobject3d} datasets as experimental benchmarks, as they provide complete ground-truth surface data for evaluating reconstruction performance in terms of noise and surface completeness. For NeRF Synthetic, we select five synthetic object scenes, each scene containing 100 training images and 200 test images, with a spatial resolution of $800 \times 800$ pixels per image. 
The OmniObject3D dataset comprises a large collection of real-world scanned objects and has been adopted in recent work, such as DiMeR~\cite{Jiang2025DiMeRDM} and SIMECO~\cite{wang2026learning}, to evaluate the quality of geometric reconstruction. Following the original partitioning scheme, we select 12 representative objects from the three difficulty levels (easy, medium, and hard) for evaluation. Each object provides training images at a resolution of $800 \times 800$. 

We evaluate the quality of the generated geometry using Chamfer Distance (CD). In particular, we sample from the complete mesh during evaluation, without neglecting challenging regions such as those with limited viewpoints. Additionally, we statistically analyze the triangle mesh quality, including maximum angle, minimum angle, radius ratio, aspect ratio, and the percentage of sliver triangles. We refer to the evaluation scripts and metrics provided by GOF~\cite{yu2024gaussian} and FlexiCubes~\cite{shen2023flexible}.

\subsection{OmniObject3D Dataset}
We utilize real scanned objects from the OmniObject3D~\cite{wu2023omniobject3d} dataset to evaluate the overall performance of our method, including reconstruction accuracy and geometric completeness.

\begin{figure*}[!t]
  \centering
  \includegraphics[width=1.0\linewidth]{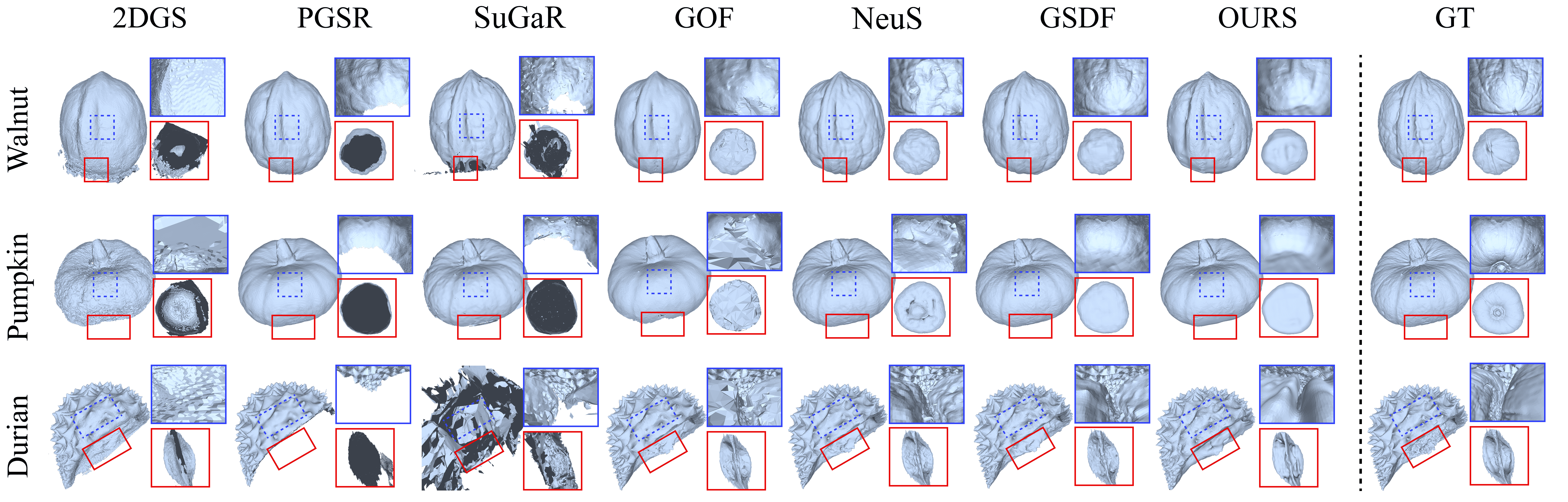} 
  \caption{Real scanned results on OmniObject3D. Interior (blue) and viewpoint‑missing bottom regions (red) are shown. Our method preserves geometric completeness while removing floaters.} 
  \label{fig:OmniObject3D_visual}
\end{figure*}

\textbf{Reconstruction Accuracy:}
The CD in Table~\ref{tab:Omini_results} indicates that our method achieves the best performance across the vast majority of scenes. Figure~\ref{fig:OmniObject3D_visual} visualizes the results. NeRF-based methods often fail to produce clean surfaces, especially for large volumetric objects. Floaters appear inside and outside the object during extraction, e.g., in walnut or pumpkin hollows (see blue rectangle). Notably, NeuS and GSDF (built on the NeuS framework) are unstable during SDF training, leading to reconstruction failures for a few objects. Gaussian-based methods produce fewer floaters but exhibit surface incompleteness, primarily because they struggle to handle object surfaces from limited viewpoints or with low-quality Gaussians. Although GOF~\cite{yu2024gaussian} achieves a lower CD, it produces cluttered triangle meshes. For methods with large errors, such as SuGar~\cite{guedon2024sugar} and 2DGS~\cite{huang20242d}, the reconstructed meshes exhibit numerous outliers scattered around the main model, resulting in incorrect geometry. This is due to their suboptimal handling of datasets with a single object and masked backgrounds.

\begin{table}[t]
  \centering
  \caption{Quantitative results on the NeRF Synthetic dataset. For resolution-controllable methods, we test two resolutions: low (128 or original lowest) and high (method default), with resolution marked by method subscript.}  
  \resizebox{\linewidth}{!}{
  \begin{tabular}{llcccccc}
  \toprule  
  CD ($10^{-2}$) $\downarrow$ & Method & Chair & Hotdog & Lego & Drums & Mic & Mean \\ 
  \midrule
  \multirow{2}{*}{Traditional} 
  & Nerf$_{128}$ & 3.62 & 4.19 & 3.07 & 6.91 & 8.31 & 5.22 \\
  & Nerf$_{512}$ & 2.16 & 2.04 & 1.85 & 1.89 & 1.24 & 1.84 \\
  \cmidrule{1-8}
  \multirow{3}{*}{SDF-NeRF}
  & VolSDF & \cellcolor{third}1.18 & 3.22 & 2.26 & 4.03 & 1.14 & 2.37 \\
  & NeuS$_{128}$ & 1.34 & 2.12 & 2.56 & 4.10 & \cellcolor{third}1.01 & 2.23 \\
  & NeuS$_{512}$ & 1.62 & \cellcolor{third}2.02 & 1.58 & 3.79 & \cellcolor{best}0.78 & 1.96 \\
  \cmidrule{1-8}
  \multirow{7}{*}{GS-based}
  & RelightableGaussian & 3.65 & 3.11 & 1.63 & 2.34 & 1.13 & 2.37 \\
  & SuGar$_{\text{Low}}$ & 2.37 & 4.01 & \cellcolor{third}1.46 & 6.21 & 4.80 & 3.77 \\
  & SuGar$_{\text{High}}$ & 1.74 & 3.18 & 1.62 & 4.86 & 4.47 & 3.17 \\
  & 2DGS$_{128}$ & 8.07 & 4.37 & 4.18 & 5.75 & 4.92 & 5.46 \\
  & 2DGS$_{1024}$ & 1.71 & 2.87 & 4.18 & 4.01 & 2.51 & 3.06 \\
  & PGSR & 1.76 & 3.14 & 1.99 & 1.54 & 1.17 & 1.92 \\
  & GOF & 2.51 & 3.63 & 2.22 & \cellcolor{third}1.42 & \cellcolor{second}0.99 & 2.15 \\
  \cmidrule{1-8}
  \multirow{3}{*}{GS-guided SDF}
  & GSDF$_{128}$ & 1.78 & 2.45 & 2.74 & 4.11 & 7.35 & 3.69 \\
  & GSDF$_{512}$ & 1.62 & 3.08 & 1.58 & 3.79 & 7.13 & 3.44 \\ 
  & 3DGSR & \cellcolor{second}1.01 & \cellcolor{best}1.48 & \cellcolor{second}1.45 & \cellcolor{second}0.95 & 1.15 & 1.21 \\
  & Ours$_{128}$ & \cellcolor{best}0.71 & \cellcolor{second}1.76 & \cellcolor{best}1.29 & \cellcolor{best}0.93 & 1.12 & \cellcolor{best}1.16 \\
  \bottomrule
  \end{tabular}
  }
  \label{tab:blender_results}  
\end{table}
\textbf{Geometric Completeness:}
Figure~\ref{fig:OmniObject3D_visual} also shows that our method preserves geometric continuity and completeness on the OmniObject3D dataset, especially in regions with missing viewpoints. This is because optimizing a continuous field enables indirect supervision of unobserved regions via images from other viewpoints. TSDF fusion fails to reconstruct geometry in areas with missing views; GOF merely fills the voids, but it struggles to follow the original shape.

\subsection{NeRF Synthetic Dataset}
\begin{figure}[!t]
\centering
\includegraphics[width=0.9\columnwidth]{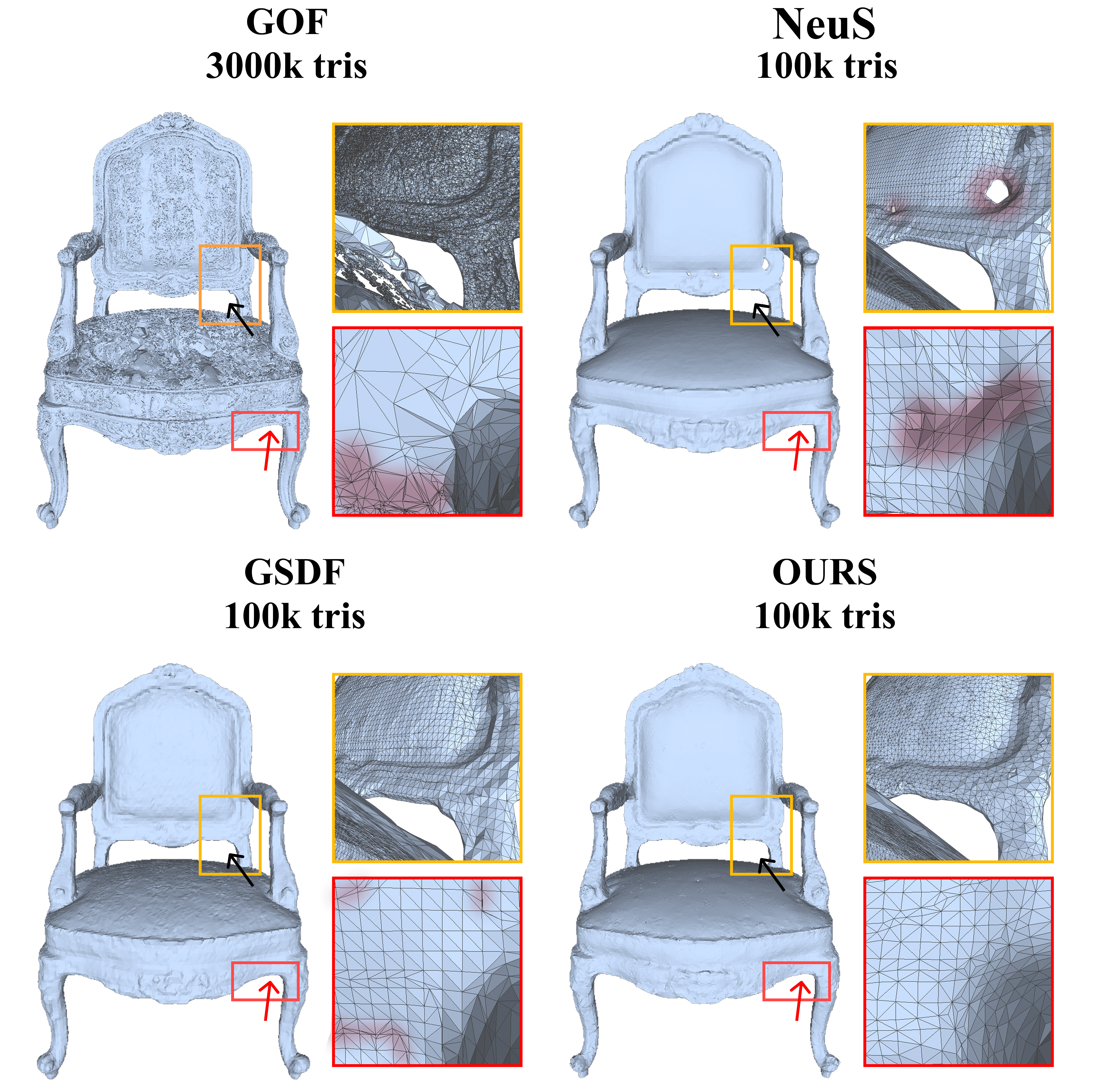}
\caption{Mesh quality. We show magnified surface details for each surface in the direction of the arrow. Regions with geometric errors and sliver triangles are highlighted in red. Our method consistently achieves the best mesh quality as optimization progresses.}
\label{fig:mesh_quality}
\vspace{-0.05cm}
\end{figure}
We utilize objects with diverse textures and fine structures from the NeRF Synthetic dataset for detailed evaluation. We focus on the performance of our method at different resolutions and the quality of the triangulated surfaces.

\textbf{Different Resolutions:} As shown in Table~\ref{tab:blender_results}, our method achieves the best average CD among all baselines. Compared with other methods, our approach consistently reconstructs high-quality surfaces even at lower resolutions. At resolutions below 128, NeRF introduces significant noise due to the difficulty of threshold selection for surface extraction, and Gaussian-based methods fail to remove redundant surfaces, thereby hindering the recovery of fine geometric details. For a few objects, our reconstruction error is slightly higher, primarily due to overly fine structures or surface reflections(e.g., drum hardware and  mic grille.) that hinder the field modeling and isosurface extraction process.

\begin{figure*}[!t]
  \centering
  \includegraphics[width=1.0\linewidth]{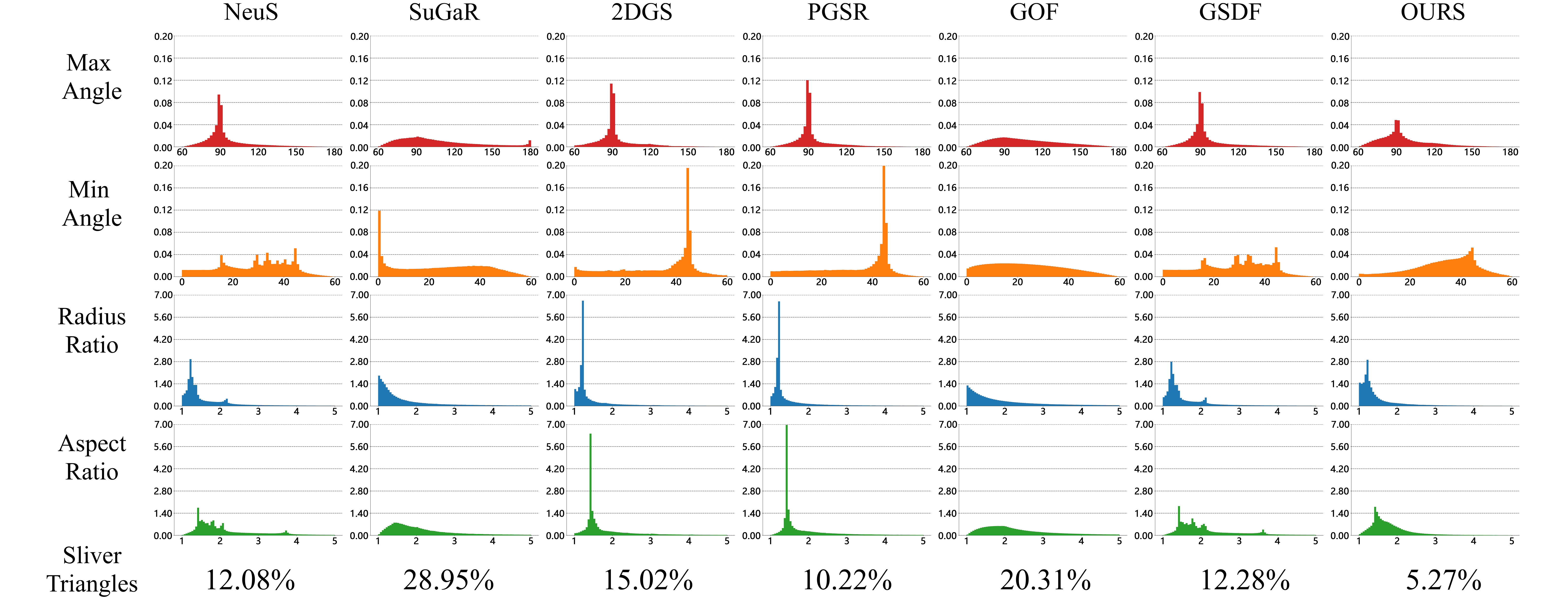} 
  \caption{Quantitative comparison of the intrinsic quality of extracted meshes. Our method yields the lowest number of sliver triangles with interior angles below $10^\circ$ and more equilateral shapes. Smoother angle distributions show adaptation to geometric variations along curved boundaries and fine details. This result highlights the need for end-to-end surface reconstruction.} 
  \label{fig:mesh}
\end{figure*}

\textbf{Mesh Quality:} Figure~\ref{fig:mesh_quality} and Figure~\ref{fig:mesh} illustrate the qualitative and quantitative comparison results of triangle mesh quality. Results show that GOF~\cite{yu2024gaussian} and SuGar~\cite{guedon2024sugar} produce cluttered, irregular triangle distributions, 2DGS~\cite{huang20242d} and PGSR~\cite{chen2024pgsr} over-concentrate on right triangles and lack geometric adaptability. NeuS~\cite{wang2021neus} and GSDF~\cite{yu2024gsdf} are prone to unstable holes and bumps. Our method generates meshes with a drastically lower fraction of sliver triangles and more regular shapes compared to all baseline approaches. The shape of triangular faces adheres more closely to the equilateral principle of Delaunay triangulation~\cite{cheng2013delaunay}.

\subsection{Ablation Study}
\begin{figure}[!t]
\centering
\includegraphics[width=1.0\columnwidth]{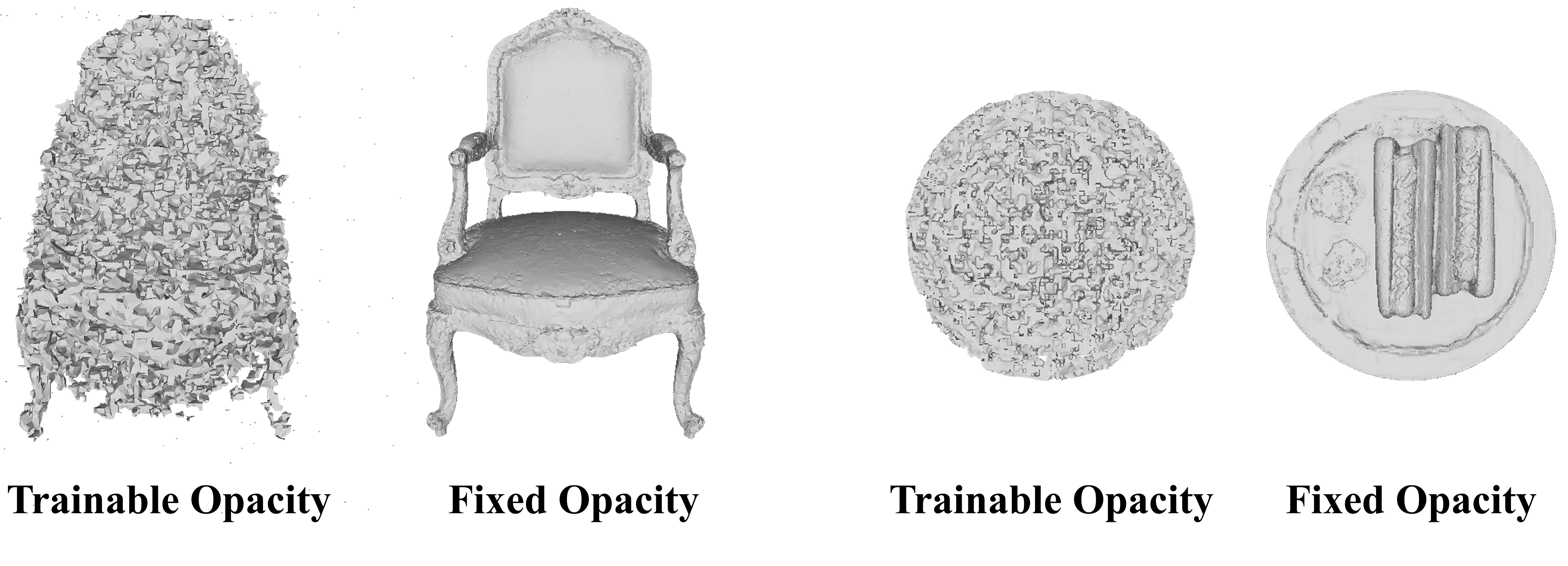}
\caption{Without the opacity constraint, low-opacity Gaussians accumulate around the surface, producing layered floating artifacts and bloated geometry inconsistent with the rendered appearance.}
\label{fig:ablation_op}
\end{figure}

\begin{figure}[!t]
\centering
\includegraphics[width=1.0\columnwidth]{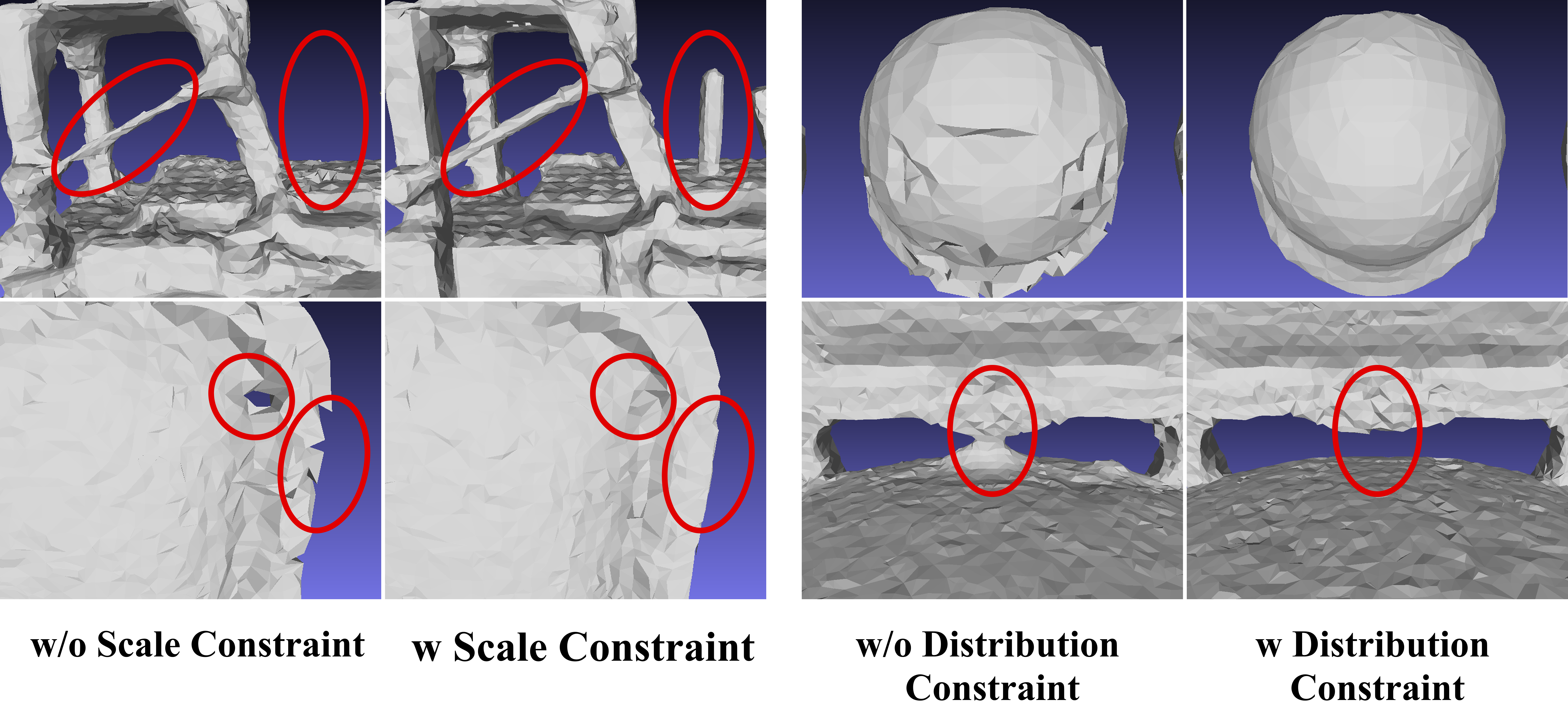}
\caption{Comparison of the effects of the scale constraint (left) and distribution constraint (right) on Chair, Lego, and Materials. Without the scale constraint, the reconstruction exhibits missing geometry. Without the distribution constraint, the surface contains irregular geometric fluctuations.}
\label{fig:ablation_sd}
\end{figure}

\begin{figure}[!t]
\centering
\includegraphics[width=1.0\columnwidth]{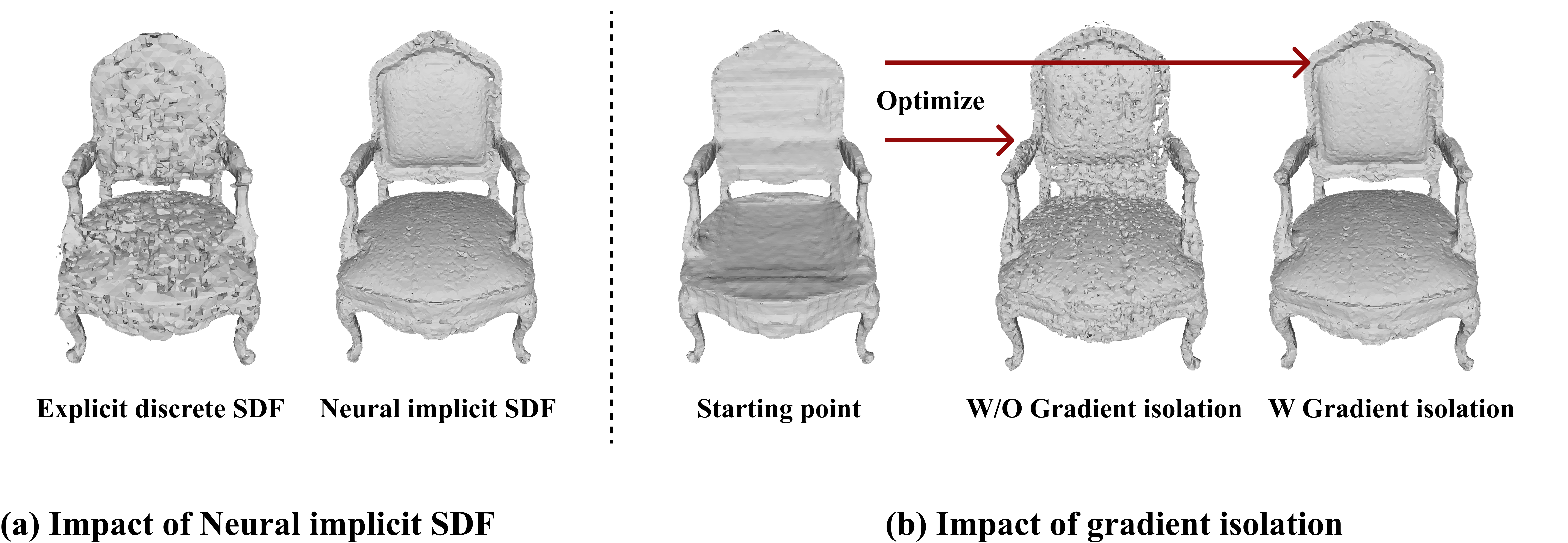}
\caption{(a) Neural implicit SDF yields smoother and more continuous surfaces, while discrete SDFs tend to produce noise.
(b) Starting from the same initialization, training without gradient isolation leads to mutual interference between surface and Gaussian optimization, resulting in degraded surface quality.
}
\label{fig:ablation_sdf}
\end{figure}

\begin{table}[!t]
\centering
\caption{Ablation Studies on the NeRF Synthetic Dataset.}
\label{tab:ablation}
\begin{tabular}{l|ccc|c}
\hline
Settings & Opacity & Scale & Distribution & CD $\downarrow$ \\
\hline
Baseline &  &  &  & 11.40 \\
Only Opacity &\checkmark  &  &  & 2.87 \\
w/o Scale & \checkmark &  & \checkmark & 1.58 \\
w/o Distribution & \checkmark & \checkmark &  &1.36  \\
Full Model & \checkmark & \checkmark & \checkmark & 1.16 \\
\hline
\end{tabular}
\end{table}

\textbf{Gaussian Constraints:} We validate the effectiveness of our Surface Gaussian Constraints on the NeRF Synthetic dataset. The results are summarized in Table~\ref{tab:ablation}.

I) \textit{Opacity}. In Sec. 3.1, we apply an opacity constraint to fix Gaussian opacity to its maximum value. As shown in Fig.~\ref{fig:ablation_op}, we show that this constraint is critical for stable geometry optimization.
Without it, erroneous regions can be represented by low-opacity Gaussians that contribute little to the rendered image. As a result, these regions receive insufficient supervision during geometric refinement and persist as floating or unstable surface artifacts.

II) \textit{Scale and Distribution}. Since scale and distribution constraints provide limited benefit on severely corrupted geometry (e.g., the Trainable Opacity mesh in Fig.~\ref{fig:ablation_op}), we evaluate these two constraints with the opacity constraint enabled. As shown in Fig.~\ref{fig:ablation_sd}, removing these constraints degrades reconstruction accuracy. This is because the spatial coverage of Gaussians no longer aligns with the underlying surface geometry. Over-expanded or incomplete Gaussian regions produce incorrect rendering supervision, leading to inaccurate surface optimization and degraded mesh quality.

\textbf{Optimization:} We qualitatively verified the improvements we made to the optimization process, and the results are summarized in Fig.~\ref{fig:ablation_sdf}.

I) \textit{Neural Implicit SDF}. The original differentiable extraction method directly optimizes discrete SDF values on voxel vertices. In contrast, we parameterize the SDF with an implicit neural network. As shown in Fig.~\ref{fig:ablation_sdf}, the neural implicit SDF produces smoother surfaces with fewer noise.

II) \textit{Gradient isolation}. In Section 3.3, we employ a gradient detachment strategy, where a detached copy of the Gaussians is optimized independently from the SDF. As shown in Fig.~\ref{fig:ablation_sdf}, this prevents gradient contamination and avoids propagating erroneous supervisory signals during training.
.

\section{Limitations}
Our method achieves strong surface quality and geometric completeness, but several limitations remain. First, the bi‑level optimization leads to long reconstruction time and heavy GPU memory usage, restricting our framework to object‑level scenes. Second, Gaussian‑based rendering remains brittle on reflective materials and fine‑scale textures where geometry cues are ambiguous. Future work will prioritize reducing runtime and GPU memory overhead by retaining Gaussians that require no re-optimization. Additionally, rendering supervision can be improved by incorporating more expressive Gaussian formulations and stronger geometric constraints to better preserve detailed surface structures.

\section{Conclusion}

We introduced \textbf{Gaussian Sculpting}, a fully differentiable framework for end-to-end surface reconstruction via SDF optimization and constrained Gaussian rendering. Our method establishes stronger consistency between rendered views and surface geometry through Gaussian constraints and bi-level optimization, enabling stable geometry refinement and reducing artifacts such as floaters, missing regions, and geometric inconsistencies. Experimental results demonstrate that our framework produces complete and more topologically regular meshes than prior NeRF-based and Gaussian-based reconstruction approaches across diverse reconstruction scenarios.

\bibliographystyle{ACM-Reference-Format}
\bibliography{sample-base}
\clearpage

\appendix

\end{document}